\pdfoutput=1

\documentclass[11pt]{article}

\usepackage[final]{acl}
\usepackage{graphicx}
\usepackage{times}
\usepackage{latexsym}
\usepackage{float}
\usepackage{booktabs}
\usepackage{multirow}
\usepackage{makecell}
\usepackage{fontawesome}
\usepackage{amsmath}
\usepackage{multirow}
\usepackage{booktabs}
\usepackage{graphicx}
\usepackage{xcolor}
\usepackage{color}
\usepackage{amsmath} 
\usepackage{amsfonts} 
\usepackage{amssymb} 
\usepackage{url}

\usepackage{enumitem}
\usepackage[T1]{fontenc}

\usepackage[utf8]{inputenc}

\usepackage{microtype}

\usepackage{inconsolata}

\title{Enhancing Noise Robustness of Retrieval-Augmented Language Models with Adaptive Adversarial Training}

\author{Feiteng Fang$^{1,2}$\thanks{Equal contribution.}, Yuelin Bai$^{2}$\footnotemark[1], Shiwen Ni$^{2}$\thanks{Corresponding author.}, Min Yang$^{2}$\footnotemark[2], Xiaojun Chen$^{3}$, Ruifeng Xu$^{4}$\\
    $^{1}$University of Science and Technology of China \\ 
    $^{2}$Shenzhen Institute of Advanced Technology, Chinese Academy of Sciences\\ 
    $^{3}$Shenzhen University
    $^{4}$Harbin Institute of Technology (Shenzhen)\\
    feitengfang@mail.ustc.edu.cn, \{yl.bai, sw.ni, min.yang\}@siat.ac.cn, \\
    xjchen@szu.edu.cn, xuruifeng@hit.edu.cn
}

\begin{document}
\maketitle
\begin{abstract}
Large Language Models (LLMs) exhibit substantial capabilities yet encounter challenges, including hallucination, outdated knowledge, and untraceable reasoning processes. Retrieval-augmented generation (RAG) has emerged as a promising solution, integrating knowledge from external databases to mitigate these challenges. 
However, inappropriate retrieved passages can potentially hinder the LLMs' capacity to generate comprehensive and high-quality responses. 
Prior RAG studies on the robustness of retrieval noises often confine themselves to a limited set of noise types, deviating from real-world retrieval environments and limiting practical applicability. 
In this study, we initially investigate retrieval noises and categorize them into three distinct types, reflecting real-world environments. We analyze the impact of these various retrieval noises on the robustness of LLMs. Subsequently, we propose a novel RAG approach known as Retrieval-augmented Adaptive Adversarial Training (RAAT). RAAT leverages adaptive adversarial training to dynamically adjust the model's training process in response to retrieval noises. Concurrently, it employs multi-task learning to ensure the model's capacity to internally recognize noisy contexts. Extensive experiments demonstrate that the LLaMA-2 7B model trained using RAAT exhibits significant improvements in F1 and EM scores under diverse noise conditions. For reproducibility,  we release our code and data at: \url{https://github.com/calubkk/RAAT}.
\end{abstract}

\section{Introduction}
\begin{figure}[h]
  \centering
  \includegraphics[width=\linewidth]{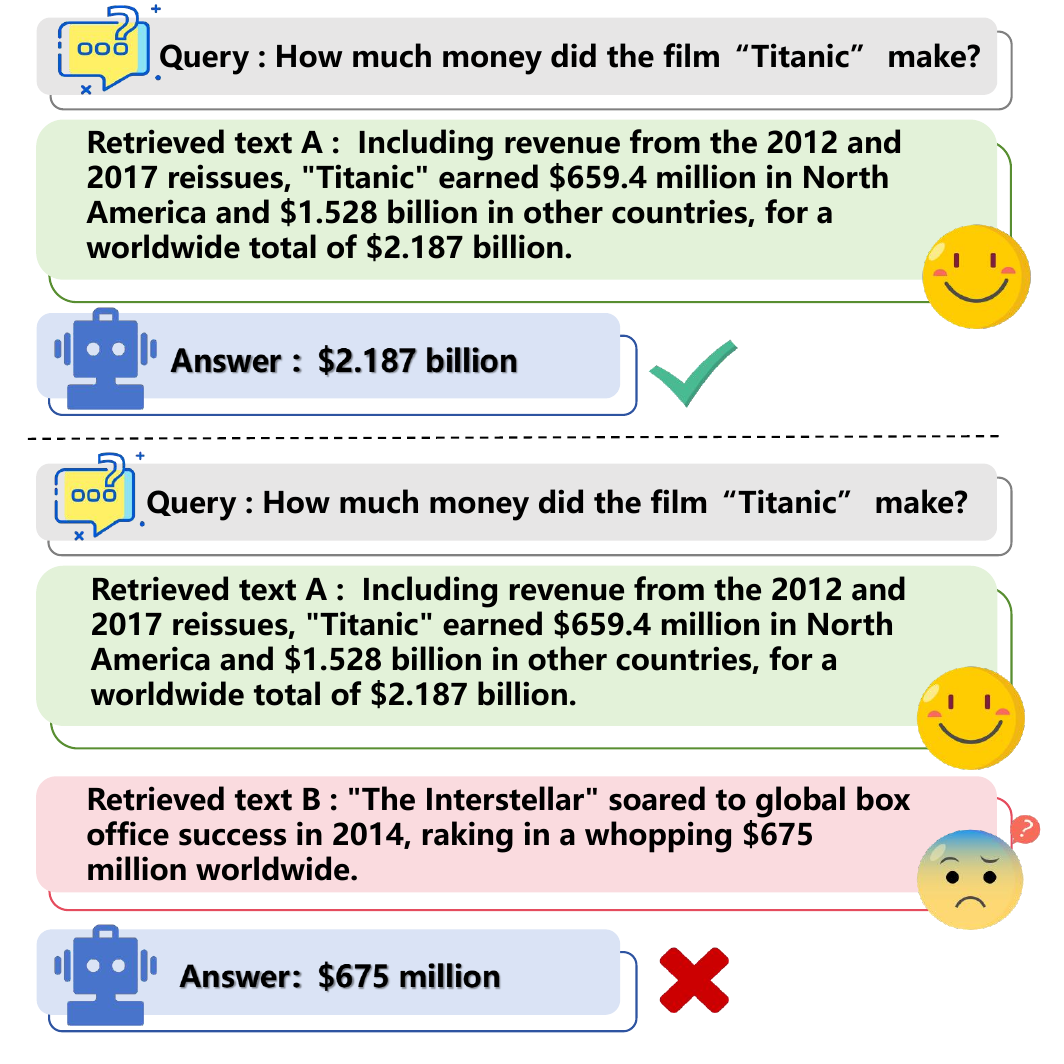}
  \caption{
  An illustrative example of the RAG process applied to question answering. The model predicts the correct answer with accurate retrieved text. However, it fails to produce the right answer when the retrieved text contains misleading or inaccurate information. 
  }
  \label{fig:display}
\end{figure}

Large language models (LLMs) have garnered substantial attention in both academic and industrial research within the domain of artificial intelligence due to their remarkable capabilities \citep{brown2020language, bubeck2023sparks}. Despite their immense power, LLMs face challenges such as hallucinations and outdated knowledge \citep{gao2023retrieval}. Moreover, a lack of domain knowledge may hinder their performance on domain-specific tasks \citep{kandpal2023large}.
To mitigate these challenges, recent studies improve LLMs by retrieving passages from external databases and pretending them in context, constituting a framework known as retrieval-augmented language models (RALMs) \citep{mao2020generation, lewis2020retrieval}.

However, RALMs also present significant limitations. Previous studies \citep{yoran2023making, yu2023chain, shi2023large} have empirically demonstrated that retrieved noisy passages are problematic for LLMs, resulting in performance degradation. We term this issue as the noise robustness problem of RALMs. As illustrated in Figure~\ref{fig:display}, the model can provide correct answers when the retrieving context is accurate and related to the query. However, when the retrieved context contains misleading or inaccurate information, the model may yield incorrect answers. As the retriever inherently cannot achieve complete accuracy, the presence of noise in the retrieved context is inevitable. 
Therefore, designing robust algorithms against retrieved noises is of great practical importance. 

Recently, several studies~\citep{yoran2023making,li2022large} have attempted to enhance the noise robustness of RALMs through noisy training, which involves incorporating retrieved noisy contexts into fine-tuning data. While noisy training exhibits promise, its effectiveness heavily relies on the composition of the training dataset. Incorrectly introducing noises to the training data can lead to model overfitting, adversely affecting generalization.
In practice, meticulous adjustment of the type and intensity of noises is essential to ensure the model's proficiency across various tasks and datasets. This demands significant experimentation and tuning, adding complexity to the development process. Moreover, the lack of clear classification for retrieval noises in current studies stands in contrast to the diverse range of noises encountered in real retrieval environments.


This paper systematically explores three types of retrieval noises: (i) contexts that are superficially related to the query but lack the correct answer (\textit{Relevant retrieval noise}), (ii) contexts that are irrelevant to the query (\textit{Irrelevant retrieval noise}), and (iii) contexts that are topically related to the query but contain incorrect information (\textit{Counterfactual retrieval noise}). Our empirical study indicates that LLMs exhibit varying robustness to these three types of noise. Compared to entirely irrelevant texts, texts that are superficially related to the query or those containing counterfactual details often lead to more misinformation. 

In response to diverse types of noises, we propose a novel approach named Retrieval-augmented Adaptive Adversarial Training (RAAT), which employs adaptive adversarial training to dynamically regulate the model's training process in response to retrieved noisy texts.
Concretely, RAAT generates adversarial samples (noises) by considering the model's sensitivity to different types of noises during training, which aligns with the min-max paradigm of adversarial training~\citep{morris2020textattack,ivgi2021achieving}. Moreover, RAAT utilizes multi-task learning~\citep{ruder2017overview} to encourage the LLMs to generate tokens that are aware of noises, thereby enabling the model to internally recognize retrieved noisy contexts and improve the overall generation performance.


The main contributions of this paper can be summarized as follows:
\begin{itemize}[leftmargin=*]
\item We systematically explore three types of retrieval noises and investigate the sensitivity of LLMs to these diverse types of noises.
\item We propose a novel adaptive adversarial training method (called RAAT) to enhance the robustness of RALMs against various retrieval noises. RAAT dynamically adjusts the training process of the model in diverse noise environments. In addition, it integrates multi-task learning to encourage the model to improve its ability to discern different types of noises. 
\item We set up a benchmark (named RAG-Bench) for assessing the noise robustness problem of RALMs based on three open-domain question-answering datasets. Experimental results demonstrate that our RAAT method enhances robustness across diverse retrieval noise environments.
\end{itemize}

\section{Related Work}
\paragraph{Retrieval-Augmented Generation with Noisy Context}
Retrieval-Augmented Language Models (RALMs) have shown impressive performance in various NLP tasks~\citep{gao2023retrieval, zhu2023large}. However, limited by the capabilities of the retriever, retrieval-augmented systems inevitably introduce irrelevant or partially relevant knowledge to the models \citep{yin2023alcuna}. Recent studies \citep{yu2023chain, yoran2023making, chen2023benchmarking} have increasingly focused on the impact of noisy information on retrieval-augmented generation. For example, \citet{jia2017adversarial, creswell2022selection}  observed that adding irrelevant noise to the context could detrimentally affect model performance. \citet{chen2023benchmarking}  demonstrated that as the proportion of noise in the retrieval context increases, the performance of LLMs experiences a notable decline.
Similar phenomena have been reported by ~\citet{yoran2023making} and ~\citet{thakur2023nomiracl}. 

\paragraph{Adversarial Training}
Adversarial training is recognized as a crucial method for enhancing model robustness, initially proposed by~\citet{goodfellow2014explaining}. 
Early studies widely investigated adversarial training in the computer vision domain~\citep{kurakin2016adversarial, madry2017towards}. In the NLP domain, ~\citet{miyato2016adversarial} applied perturbations to word embeddings, making the model less prone to overfitting. 
Similarly, perturbations on different granularities have been extensively studied, encompassing various aspects of NLP tasks~\citep{yasunaga2017robust, wu2017adversarial, zhu2019freelb, wang2020infobert,ni2023dropattack,liang2023knowledge}.  


Recently, several studies have concentrated on generating adversarial examples designed to induce LLMs to generate harmful or non-factual content \citep{zou2023universal,shen2023chatgpt} instead of merely causing the model to make inaccurate predictions.
\citet{shen2023chatgpt} employed decision-based perturbation at different levels to craft adversarial examples, revealing vulnerabilities in ChatGPT to both sentence-level and character-level adversarial attacks.
\citet{shi2023large} added irrelevant context to an arithmetic reasoning dataset, finding that including irrelevant information distracted the model's predictions. 
\citet{zou2023universal} proposed a method that could reliably generate adversarial attack suffixes, yielding adversarial prompts that exhibit high transferability. 

In this study, we investigate adversarial training concerning LLMs in response to various retrieval noises, aiming to efficiently obtain adversarial examples that enhance model robustness while reducing training overhead. We construct noisy adversarial examples by sampling or paraphrasing the original dataset. 
This approach ensures more dependable and precise outputs even when confronted with imperfect retrieved contexts.


\section{Methodology}
\subsection{Problem Setup}
In the standard RALM, given input query $x$, a retriever $r$ is designed to retrieve relevant contexts $C=\{c_1, c_2, \ldots\}$ from an external database. During inference, the content of retrieval context is concatenated with $x$ to form $d$, which is then fed into the pre-trained language model $M$, yielding a response $\hat{y}$ regarding $x$. If the retrieved context $c$ contains the correct answer $y$ about $x$, we can denote $c$ as $c_{golden}$, representing the golden retrieval context. However, if $c$ does not contain the correct answer $y$ or contains partially irrelevant content, we can denote $c$ as $c_{noisy}$.

In our study, we transform open-domain question answering (QA) into a reading comprehension task to meet the RAG settings. Formally, given the objective $f$ of an open domain question answering task is $f:\left\{x\right\} \rightarrow y$, we can formulate the objective of the reading comprehension task as $f:\left\{c_{golden}, x\right\} \rightarrow y$.
In examining the challenge of the retrieval noise robustness problem of RALM, we aim to obtain a fine-tuned model $M^{\prime}$ that can not only fulfill the function $f:\left\{c_{golden}, x\right\} \rightarrow y$ but also produce accurate answers even in the presence of additional retrieval noise $c_{noisy}$ and achieve function 
$f:\left\{c_{golden}, c_{noisy}, x\right\} \rightarrow y$.


\begin{figure*}[h]
  \centering
  \includegraphics[width=\linewidth]{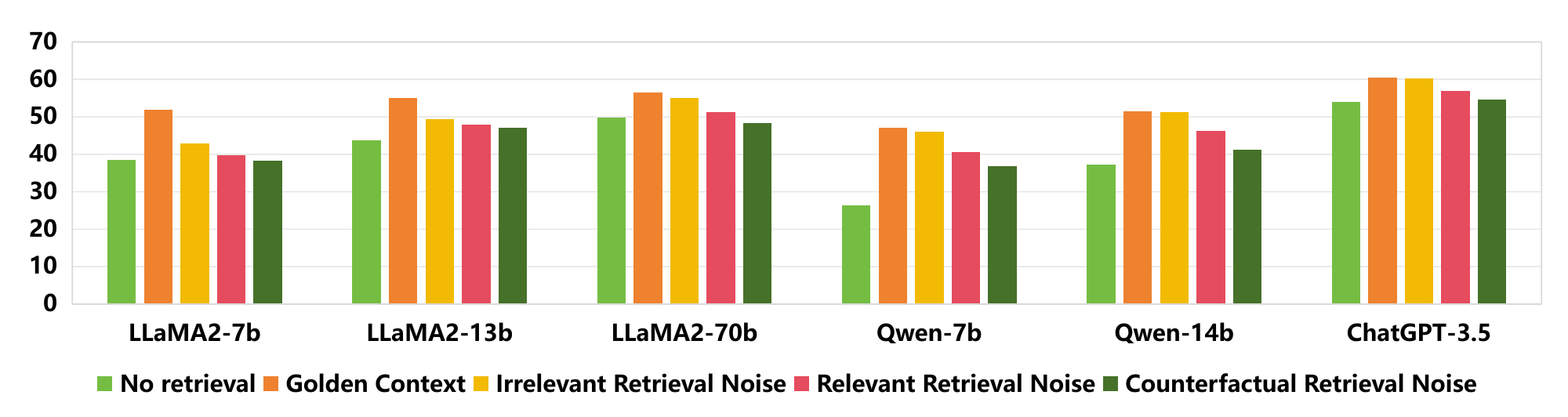}
  \caption{Exact match (EM) scores of various models under different types of retrieval noises. ``\textit{Golden Context}'' denotes instances where LLMs respond to questions with reference to the golden retrieval context. ``\textit{No Noise}'' indicates instances where LLMs answer questions without any retrieval.  The experimental configurations of other models involve the introduction of different types of noises on the foundation of the ``\textit{Golden Context}''.}
  \label{fig:empirical}
\end{figure*}

\subsection{Diverse Retrieval Noises}
We systematically classify the retrieval noise present in $c_{noisy}$ to closely mimic real-world conditions. Existing studies~\citep{yoran2023making,yu2023chain} on retrieval noise robustness often dichotomize noise into relevant and irrelevant categories. However, we contend that such a classification may not fully align with the retrieval noise robustness of RALMs.
In this work, we propose a more nuanced classification of retrieval noise, differentiating it into three distinct types: \textit{Relevant retrieval noise}, \textit{Irrelevant retrieval noise}, and \textit{Counterfactual retrieval noise}.
Specifically, \textit{Relevant retrieval noise} (denoted as $c_r$) pertains to contexts that exhibit superficial relevance to the query $x$ but lack the information necessary for the correct answer $y$. These contexts may appear relevant at first glance but ultimately mislead the model. \textit{Irrelevant retrieval noise} (denoted as $c_i$) encompasses contexts with low relevance to the query $x$, often arising from erroneous retrievals and generally being off-topic. \textit{Counterfactual retrieval noise} (denoted as $c_c$) encompasses contexts that are topically related to $x$ but contain incorrect and misleading information, often attributed to inaccuracies in the retriever's database.


To examine the influence of three distinct types of retrieval noise on LLMs, we establish a benchmark for
assessing retrieval noise robustness in LLMs by employing three open-domain question-answering datasets: Natural Questions~\citep{kwiatkowski2019natural}, TriviaQA~\citep{joshi2017triviaqa}, and WebQ~\citep{berant2013semantic}. Leveraging this benchmark, we evaluated the susceptibility of various open-source large language models to the effects of the three identified types of noise. The details of the construction of this benchmark can be found in Section \ref{sec:data}. Leveraging this benchmark, we evaluate the sensitivity of various LLMs to the effects of the three types of noise. Specifically, we conduct experiments on six LLMs, including ChatGPT$_{3.5}$, LLaMA2$_{7B}$~\citep{touvron2023llama}, LLaMA2$_{13B}$~\citep{touvron2023llama}, LLaMA2$_{70B}$~\citep{touvron2023llama}, Qwen$_{7B}$~\citep{bai2023qwen}, and Qwen$_{14B}$~\citep{bai2023qwen}.
For each model, our experiments encompass two distinct settings: one with the exclusive presence of the golden retrieval context $c_{golden}$ and another incorporating the introduction of three different types of retrieval noise $c_{noisy}$.
As shown in Figure~\ref{fig:empirical}, all LLMs experience varying degrees of impact from the three types of noise. The performance of LLMs exhibits a decline ranging from 0.2\% to 13.43\%. Through a comparative analysis of the effects of the three types of noise, we observe that \text{irrelevant retrieval noise} has a comparatively minor impact on LLMs with substantial capabilities.

\begin{figure*}[h]
  \centering
  \includegraphics[width=\linewidth]{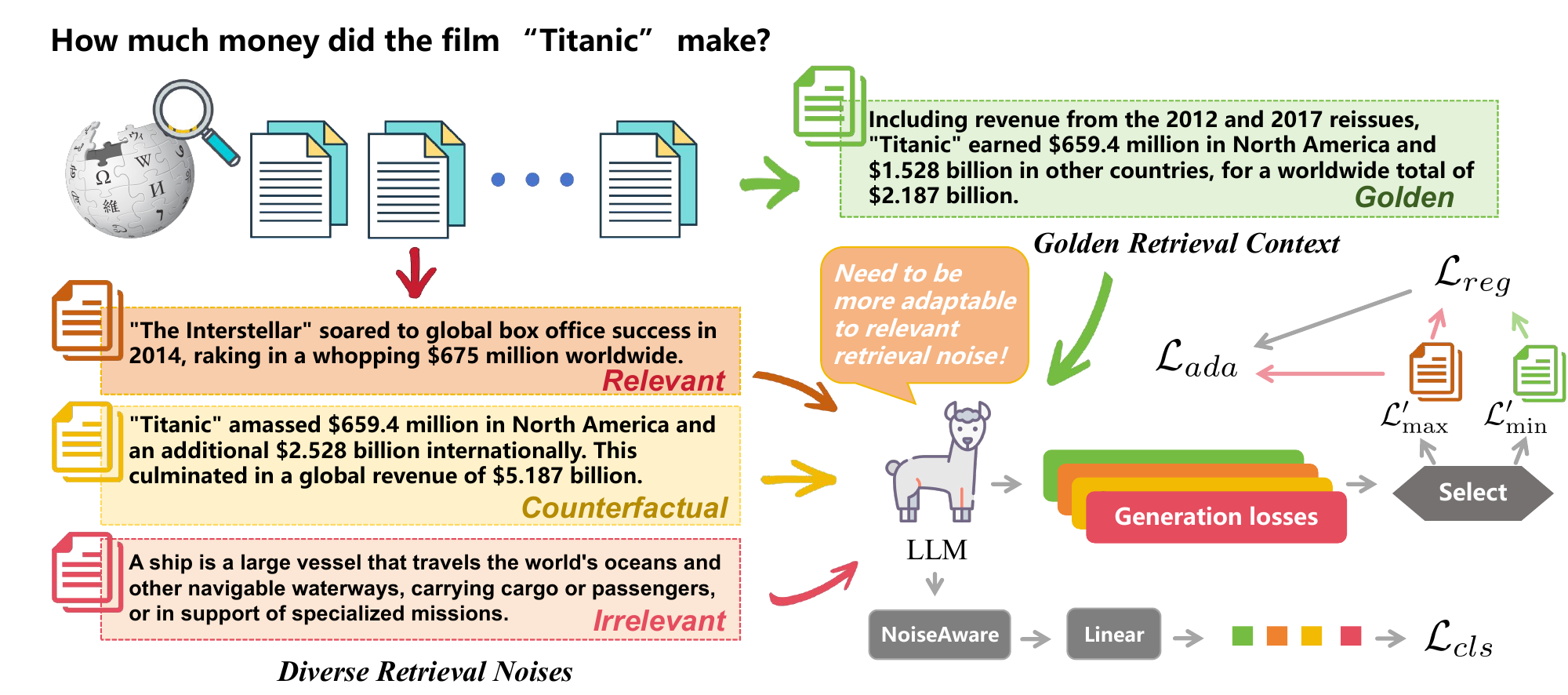}
  \caption{The overview of our proposed RAAT method, which incorporates three distinct types of retrieval noises and the golden retrieval context during the training process.}
  %
  \label{fig:overview}
\end{figure*}

\subsection{Retrieval-augmented Adaptive Adversarial Training}
Recently, several studies~\citep{yoran2023making,li2022large} attempted to enhance the noise robustness of LLMs through noisy training, which involves incorporating retrieved noisy contexts into fine-tuning data. The essence of noisy training involves the exploration of offline data augmentation, while in contrast, adversarial training leverages online data augmentation for a similar purpose~\citep{ivgi2021achieving}. The core idea of adversarial training is to fortify the models against adversarial conditions by introducing adversarial perturbations~\citep{jain2023neftune}. 
In the construction of adversarial samples, also known as noise samples, the min-max optimization strategy assumes a pivotal role, encompassing two fundamental steps. Initially, the maximization process involves adjusting the input data to intentionally mislead the model, inducing the maximum prediction error. Then, the minimization process entails fine-tuning the model's parameters to enhance its resistance against these meticulously crafted input perturbations~\citep{bai2021recent}. This strategy seeks to strike a balance, allowing the model to accurately identify normal data while robustly defending against potential attacks from adversarial examples.

In this study, we aim to refine the objective of adversarial training while exploring the noise robustness challenges of RALMs.
Considering a given query $x$, we assume the existence of four types of data augmentation, namely, golden retrieval context only ($da_g$), additional relevant retrieval noise ($da_r$), additional irrelevant retrieval noise ($da_i$), and additional counterfactual retrieval noise ($da_c$). The space of data augmentation is denoted as ${DA}=\{{da}_g,{da}_r,{da}_i,{da}_c\}$. Then, the optimization problem can be formulated as follows:
\begin{equation}
\min _\theta \mathbb{E}_{(x, y) \sim \mathcal{D}}\left[\max _{da \in DA} \mathcal{L}(\theta, da(x) , y)\right]
\end{equation}
where $\mathcal{D}$ denotes training data,  $\mathcal{L}$ is the loss function, $\theta$ denotes the parameters of LLMs, and $da(x)$ represents the data augmentation of $x$.  

Building upon the optimization problem outlined above, we introduce adaptive adversarial training as a tailored approach to enhance the robustness of RALMs against retrieval noise. Within adaptive adversarial training, the model refrains from updating parameters across all adversarial samples. Instead, it initiates the process by computing the generation loss for each adversarial sample, quantifying its adaptability to varying noise environments. Notably, a higher generation loss implies reduced adaptability of the model to the noisy environment. Given that each query involves one sample with a golden retrieval context and three adversarial samples, the model generates four distinct generation losses in each iteration. Following a min-max optimization strategy, the model prioritizes the selection of the largest loss to guide subsequent parameter update.
Formally, we define the generation loss function $\mathcal{L}^\prime$ for the augmented input $x^\prime$ as:
\begin{equation}
\mathcal{L}^\prime\left(\theta, x^\prime, y\right)=-\frac{1}{\left|y\right|} \sum_{t=1}^{\left|y\right|} \log P_{\theta}\left(y_t \mid x^\prime, y_{<t}\right)
\end{equation}
where $x^\prime=da(x)$ represents the augmented noise context of $x$.

To effectively enhance performance across diverse noise environments, adaptive adversarial training incorporates a regularization term into its loss function. 
This incorporation of a regularization term is designed to mitigate the risk of the model overfitting to a particular type of noise. The regularization term acts as a stabilizing factor, promoting generalization and preventing the model from becoming overly specialized in its response to a specific noise profile.
To achieve this goal, we introduce a regularization term specifically designed to mitigate the variance between these generation losses. 
This regularization term operates by identifying the largest $\mathcal{L}^\prime_{max}$ and the smallest $\mathcal{L}^\prime_{min}$ of the four generation losses at each training step. $\mathcal{L}^\prime_{max}$ is the generation loss with the highest numerical value among four losses being considered. Conversely, $\mathcal{L}^\prime_{min}$ is the loss function with the lowest numerical value. Here, an increased loss value indicates a greater magnitude of error or disparity in the aspect of the model's performance being assessed. This suggests that the model exhibits heightened sensitivity to adversarial examples reflecting retrieval noise.
These adversarial examples are designed to probe and exploit weaknesses in the model's processing capabilities, especially in how it deals with noisy information in its input data. The regularization term, calculated as the square of the difference between $\mathcal{L}^\prime_{max}$ and $\mathcal{L}^\prime_{min}$, aims to reduce the model's sensitivity to retrieval noise by encouraging a more balanced optimization. Formally, we design the regularization term $\mathcal{L}_{\text{reg}}$ as: 
\begin{equation}
\mathcal{L}_{\text{reg}} = \|\mathcal{L}^\prime_{\text{max}} - \mathcal{L}^\prime_{\text{min}}\|_2^2
\end{equation}     

Subsequently, we define the adaptive adversarial training loss function $\mathcal{L}_{\text{ada}}$ as follows:
\begin{equation}
\mathcal{L}_{\text{ada}} = \mathcal{L}^\prime_{\text{max}} + w_{\text{\text{reg}}} \cdot \mathcal{L}_{\text{reg}}
\end{equation}
where $w_{\text{reg}}$ is a pre-defined hyperparameter to control the weight of $\mathcal{L}_{\text{reg}}$.

\subsection{Incorporating Noise Awareness}
Accurately identifying retrieval noise plays a pivotal role in fortifying the robustness of RALMs against the retrieval noise. Models endowed with the ability to discern different types of noise can more effectively choose and utilize training data, leading to an improvement in the overall quality of their generated outputs. This capacity to distinguish between various noise types contributes significantly to the model's adaptive learning process, enabling it to optimize performance in the presence of diverse noise scenarios.
Inspired by the above motivation, we propose an auxiliary task designed to autonomously recognize the types of noisy retrieval texts, aiming to significantly bolster the retrieval robustness of RALMs. This auxiliary task serves as a valuable augmentation, contributing to the overall adaptability and effectiveness of the model in scenarios involving retrieval noise.

Specifically, we attempt to enable the model to generate tokens that are sensitive to noise, thereby improving the model's capacity to discern various types of retrieval noise internally.
Specifically, we first incorporate a linear layer beneath LLMs.
Subsequently, a classification loss $\mathcal{L}_{\text{cls}}$ is computed for each of the golden retrieval context and the three adversarial samples corresponding to each input $x$.
One-hot encoding is employed in classification tasks, assigning values from 1 to 4 as labels to train the classifiers, where each classifier is tailored to a different retrieval noise type. The loss function $\mathcal{L}_{\text{cls}}$ is computed using cross-entropy.

Finally, we formulate the final RAAT loss $\mathcal{L}_{\text{RAAT}}$ by combining the adaptive adversarial training loss and the classification loss in the context of multi-task learning:
\begin{equation}
\mathcal{L}_{\text{RAAT}} = w_{\text{ada}} \cdot \mathcal{L}_{\text{ada}} + w_{\text{cls}} \cdot \mathcal{L}_{\text{cls}}
\end{equation}
where $w_{\text{ada}}$ and $w_{\text{cls}}$ represent pre-defined hyperparameters used to balance the importance of these two different tasks.

\section{Experiments}
\subsection{Dataset Construction}
\label{sec:data}
We have formulated a benchmark named RAG-Bench that is specifically designed to evaluate the retrieval noise robustness of LLMs. RAG-Bench is established upon three widely available datasets that center around open-domain question answering (QA): Natural Questions (NQ) \citep{kwiatkowski2019natural}, TriviaQA \citep{joshi2017triviaqa}, and WebQ~\citep{berant2013semantic}.
For each dataset, we employ the retrieval model DPR~\citep{karpukhin2020dense} as our retriever, which retrieves ten passages from Wikipedia for each query. Then, we apply filtering to the queries, ensuring that each query in the filtered subset contains at least two golden retrieval contexts, indicating the presence of correct answers. Detailed statistics for both the full set and the filtered subset can be found in Table~\ref{tab:dataset_quantities}.

Each sample in our dataset contains a golden retrieval context and is deliberately designed to incorporate three types of augmented retrieval noise. To introduce \textbf{\textit{relevant retrieval noise}}, we choose the context most pertinent to the query from the set of ten retrieval texts, excluding the golden retrieval context. In the case of \textbf{\textit{irrelevant retrieval noise}}, no selection is made from the retrieval texts associated with the current query. Instead, a passage is randomly chosen from the retrieval contents of other queries, ensuring its complete irrelevance to the current query. For the \textbf{\textit{counterfactual retrieval noise}}, we randomly select one passage from the two golden retrieval contexts and substitute its answer entity with an incorrect one.


The test set of RAG-Bench comprises 1000 randomly chosen samples from the test sets of three QA datasets, resulting in a total of 3000 samples. The training set consists of 1500 samples randomly selected from the training sets of the three datasets, totaling 4500 samples. The validation set, drawn from the training sets of three QA datasets, contains 300 samples. Notably, careful measures were taken to ensure no overlap with the training data of RAG-Bench.
\begin{table}[t]
\begin{center}
\resizebox{\columnwidth}{!}{\begin{tabular}{ccccc}
\toprule
\multirow{2}{*}{\textbf{Datasets}} & \multicolumn{2}{c}{\textbf{Train}} & \multicolumn{2}{c}{\textbf{Test}} \\
\cmidrule(lr){2-5}
& \#Full & \#Subset & \#Full & \#Subset   \\
\midrule
\text{NQ} & 79,168 & 40,551 & 3,610 & 1,833  \\
\text{TriviaQA} & 78,785 & 51,202 & 11,313 & 7,010 \\
\text{WebQ} & 3,778 & 2,316 & 2,032 & 1,057 \\
\bottomrule
\end{tabular}}
\caption{The statistics of the three QA datasets.}
\label{tab:dataset_quantities}
\end{center}
\end{table}
\begin{table*}[!ht]
\begin{center}
\resizebox{\textwidth}{!}{\begin{tabular}{ccccccccccccc}
\toprule
\multirow{2}{*}{\textbf{Method}}&\multicolumn{2}{c}{\textbf{Golden Only}}&\multicolumn{2}{c}{\textbf{Golden \& $c_i$}}&\multicolumn{2}{c}{\textbf{Golden \& $c_r$}}&\multicolumn{2}{c}{\textbf{Golden \& $c_c$}}&\multicolumn{4}{c}{\textbf{Avg}}\\
\cmidrule(lr){2-13}
&\multicolumn{1}{c}{F1}&\multicolumn{1}{c}{EM}&\multicolumn{1}{c}{F1}&\multicolumn{1}{c}{EM}&\multicolumn{1}{c}{F1}&\multicolumn{1}{c}{EM}&\multicolumn{1}{c}{F1}&\multicolumn{1}{c}{EM}&\multicolumn{2}{c}{F1}&\multicolumn{2}{c}{EM}\\
\midrule
\multirow{1}{*}{\text{LLaMA2$_{7B}$}}&\multicolumn{1}{c}{65.56}&\multicolumn{1}{c}{51.80}&\multicolumn{1}{c}{56.14}&\multicolumn{1}{c}{42.87}&\multicolumn{1}{c}{53.10}&\multicolumn{1}{c}{39.73}&\multicolumn{1}{c}{51.81}&\multicolumn{1}{c}{38.37}&\multicolumn{2}{c}{56.68}&\multicolumn{2}{c}{43.19}\\
\multirow{1}{*}{\text{Qwen$_{7B}$}}&\multicolumn{1}{c}{62.57}&\multicolumn{1}{c}{47.07}&\multicolumn{1}{c}{61.48}&\multicolumn{1}{c}{46.06}&\multicolumn{1}{c}{55.50}&\multicolumn{1}{c}{40.50}&\multicolumn{1}{c}{53.26}&\multicolumn{1}{c}{36.90}&\multicolumn{2}{c}{58.20}&\multicolumn{2}{c}{42.63}\\
\multirow{1}{*}{\text{LLaMA2$_{13B}$}}&\multicolumn{1}{c}{69.27}&\multicolumn{1}{c}{55.00}&\multicolumn{1}{c}{63.25}&\multicolumn{1}{c}{49.47}&\multicolumn{1}{c}{62.27}&\multicolumn{1}{c}{47.97}&\multicolumn{1}{c}{62.07}&\multicolumn{1}{c}{47.17}&\multicolumn{2}{c}{64.22}&\multicolumn{2}{c}{49.90}\\
\multirow{1}{*}{\text{Qwen$_{14B}$}}&\multicolumn{1}{c}{67.45}&\multicolumn{1}{c}{51.43}&\multicolumn{1}{c}{66.71}&\multicolumn{1}{c}{51.20}&\multicolumn{1}{c}{61.88}&\multicolumn{1}{c}{46.16}&\multicolumn{1}{c}{58.65}&\multicolumn{1}{c}{41.30}&\multicolumn{2}{c}{63.67}&\multicolumn{2}{c}{47.52}\\
\multirow{1}{*}{\text{LLaMA2$_{70B}$}}&\multicolumn{1}{c}{71.43}&\multicolumn{1}{c}{56.56}&\multicolumn{1}{c}{70.05}&\multicolumn{1}{c}{55.13}&\multicolumn{1}{c}{65.97}&\multicolumn{1}{c}{51.33}&\multicolumn{1}{c}{63.91}&\multicolumn{1}{c}{48.27}&\multicolumn{2}{c}{67.84}&\multicolumn{2}{c}{52.82}\\
\multirow{1}{*}{\text{ChatGPT$_{3.5}$}}&\multicolumn{1}{c}{73.98}&\multicolumn{1}{c}{60.50}&\multicolumn{1}{c}{72.24}&\multicolumn{1}{c}{60.30}&\multicolumn{1}{c}{70.65}&\multicolumn{1}{c}{56.89}&\multicolumn{1}{c}{69.00}&\multicolumn{1}{c}{54.64}&\multicolumn{2}{c}{71.47}&\multicolumn{2}{c}{58.10}\\
\midrule
 \multirow{1}{*}{RALM$_{golden}$}&\multicolumn{1}{c}{80.31}&\multicolumn{1}{c}{74.03}&\multicolumn{1}{c}{79.33}&\multicolumn{1}{c}{72.73}&\multicolumn{1}{c}{73.26}&\multicolumn{1}{c}{66.33}&\multicolumn{1}{c}{73.08}&\multicolumn{1}{c}{65.40}&\multicolumn{2}{c}{76.50}&\multicolumn{2}{c}{69.62}\\
 
\multirow{1}{*}{RetRobust}&\multicolumn{1}{c}{80.10}&\multicolumn{1}{c}{73.80}&\multicolumn{1}{c}{79.25}&\multicolumn{1}{c}{72.97}&\multicolumn{1}{c}{74.81}&\multicolumn{1}{c}{68.30}&\multicolumn{1}{c}{75.46}&\multicolumn{1}{c}{68.43}&\multicolumn{2}{c}{77.41}&\multicolumn{2}{c}{70.88}\\

\multirow{1}{*}{RALM$_{retrieved}$}&\multicolumn{1}{c}{80.04}&\multicolumn{1}{c}{73.40}&\multicolumn{1}{c}{81.09}&\multicolumn{1}{c}{74.80}&\multicolumn{1}{c}{75.99}&\multicolumn{1}{c}{69.10}&\multicolumn{1}{c}{73.10}&\multicolumn{1}{c}{65.67}&\multicolumn{2}{c}{77.55}&\multicolumn{2}{c}{70.74}\\

\multirow{1}{*}{RALM$_{multiple}$}&\multicolumn{1}{c}{85.47}&\multicolumn{1}{c}{80.17}&\multicolumn{1}{c}{85.27}&\multicolumn{1}{c}{81.20}&\multicolumn{1}{c}{83.07}&\multicolumn{1}{c}{78.33}&\multicolumn{1}{c}{83.25}&\multicolumn{1}{c}{79.23}&\multicolumn{2}{c}{84.27}&\multicolumn{2}{c}{79.73}\\

\multirow{1}{*}{\text{RAAT}}&\multicolumn{1}{c}{\textbf{87.15}}&\multicolumn{1}{c}{\textbf{83.07}}&\multicolumn{1}{c}{\textbf{86.80}}&\multicolumn{1}{c}{\textbf{82.73}}&\multicolumn{1}{c}{\textbf{85.14}}&\multicolumn{1}{c}{\textbf{81.00}}&\multicolumn{1}{c}{\textbf{86.29}}&\multicolumn{1}{c}{\textbf{82.10}}&\multicolumn{2}{c}{\textbf{86.35}}&\multicolumn{2}{c}{\textbf{82.23}}\\
\bottomrule
\end{tabular}}
\caption{Experimental results on our RAG-Bench benchmark. ``Golden Only'' denotes a scenario where LLMs only consult the golden retrieval context. 
In ``Golden \& $c_i$/$c_r$/$c_c$'', LLMs consider both the golden retrieval context and \textit{irrelevant retrieval noise}/\textit{relevant retrieval noise}/\textit{counterfactual retrieval noise}.}
\label{tab:results}
 \end{center}
\end{table*}

\subsection{Evaluation Metrics}
We evaluate the effectiveness of our method using two metrics: exact match (EM) and F1 score~\citep{chen2017reading}. Concretely, EM assesses the extent to which the answer generated by the system aligns precisely with the standard answer without any disparities at the character level. In contrast, the F1 score incorporates precision and recall, accounting for the equilibrium between correctly identifying answers and avoiding omitting correct answers. 


\subsection{Baseline Methods}
We conduct a comparison of our RAAT method against zero-shot LLMs, as well as finetuning approaches applied to LLaMA2$_{7B}$, which shares a common backbone with RAAT.
\paragraph{\textbf{Zero-Shot} Methods}
Within the open-source community, many foundational and supervised fine-tuning (SFT) models have emerged. In our experiments, we select six renowned LLMs as baselines: ChatGPT$_{3.5}$,  LLaMA2$_{7B}$~\citep{touvron2023llama}, LLaMA2$_{13B}$, LLaMA2$_{70B}$, Qwen$_{7B}$~\citep{bai2023qwen}, and Qwen$_{14B}$.


\paragraph{\textbf{Fine-tuning Methods}}
We further compare RAAT with various fine-tuning methods.
\begin{itemize}[leftmargin=*]
\item \textbf{RALM$_{golden}$} This is a RALM with instruction tuning~\citep{lin2023ra}. It prepends a golden retrieval text \(c_{golden}\) in context to fine-tune LLaMA2$_{7B}$.
\item \textbf{RetRobust} To ensure that the model is exposed to both golden retrieval texts and various retrieval noise during training, ~\citet{yoran2023making} proposes RetRobust. For each query, RetRobust selects top-1, low-ranked, or random retrieved passages with equal probability for training.
\item \textbf{RALM$_{retrieved}$} This variant is a RALM incorporating instruction tuning. In contrast to RALM$_{golden}$, RALM$_{retrieved}$ does not manually design retrieval noise in the training set but directly uses the top-2 retrieved passages. This training method is more aligned with real retrieval environments.
\item \textbf{RALM$_{multiple}$} This approach closely resembles RetRobust, differing only in the construction of the training dataset. In RALM$_{multiple}$, rather than introducing one type of retrieval noise randomly for each query, each type of retrieval noise is combined with the sample and incorporated into the dataset. That is, each query is associated with four augmented noisy samples.
\end{itemize}
\begin{table*}[h!]
\begin{center}
\resizebox{\textwidth}{!}{\begin{tabular}{ccccccccccc}
\toprule
\multirow{2}{*}{\textbf{Method}} & \multicolumn{2}{c}{\textbf{Golden Only}} & \multicolumn{2}{c}{\textbf{Golden \& $c_i$}} & \multicolumn{2}{c}{\textbf{Golden \& $c_r$}} & \multicolumn{2}{c}{\textbf{Golden \& $c_c$}} & \multicolumn{2}{c}{\textbf{Avg}} \\
\cmidrule(lr){2-11}
& F1 & EM & F1 & EM & F1 & EM & F1 & EM & F1 & EM \\
\midrule
\text{RAAT} & 87.15 & 83.07 & 86.80 & 82.73 & 85.14 & 81.00 & 86.29 & 82.10 & 86.35  & 82.23 \\
\text{RAAT w/o $\mathcal{L}_\text{cls}$} & 86.76 & 82.77 & 86.45 & 82.27 & 84.69 & 80.63 & 85.54 & 81.20 & 85.86  & 81.71 \\
\text{RAAT w/o $\mathcal{L}_\text{reg}$} & 86.87 & 83.03 & 83.92 & 79.86 & 84.69 & 80.57 & 87.02 & 82.80 & 85.63  & 81.57 \\
\bottomrule
\end{tabular}}
\caption{Ablation test results in terms of EM and F1 score.}
\label{tab:ablation}
\end{center}
\end{table*}
\subsection{Implementation Details}
Our RAAT method relies on LLaMA2-7B as the foundational model. We set the weight parameters as follows: $w_{\text{reg}}$ to 0.1, $w_{\text{raat}}$ to 2, and $w_{\text{cls}}$ to 1. The sequence length, epoch, and learning rate are configured to 512, 2, and 5e-6, respectively. Our experiments are conducted on a computational cluster equipped with 4 NVIDIA A100 GPUs, each boasting a capacity of 80GB.

\section{Experimental Results }
\subsection{Main Results}
Table~\ref{tab:results} illustrates the efficacy of our RAAT method compared to the baselines in terms of F1 and EM scores. We observe that all models are affected by three different types of retrieval noise attacks.
The influence of \textit{irrelevant retrieval noise} is marginal, while \textit{counterfactual retrieval noise} exerts the most significant impact. For the models sharing the same architecture, larger parameter sizes correlate with superior performance and better robustness against retrieval noise. For instance, LLaMA2$_{7B}$ exhibits a 12.46\% reduction in F1 score when confronted with \textit{relevant retrieval noise}, whereas LLaMA2$_{13B}$ only experiences a 7\% decrease under identical conditions. This trend is also evident in Qwen.


From Table~\ref{tab:results}, we can also observe that fine-tuning enables LLMs to better utilize information from the retrieval texts. Fine-tuned models significantly outperform the zero-shot LLMs with varying parameter sizes. Moreover, RALM$_{multiple}$ shows a significant improvement over RALM$_{golden}$, RALM$_{retrieved}$ and RetRobust, reflecting the sensitivity of retrieval noise to the training dataset and the importance of diversity in noise attacks during training.
Our RAAT method achieves even better performance than RALM$_{multiple}$ in all four environments, with an average increase of 2.08\% in the F1 score and 2.5\% in the EM score, demonstrating its superior ability to handle diverse retrieval noise.


\subsection{Ablation Study}
To gain a comprehensive understanding of the individual contribution of each component within  RAAT to the overall performance, we conducted an ablation study by removing the regularization term loss (denoted as w/o $\mathcal{L}_\text{reg}$) and the noise-aware classification loss (denoted as w/o $\mathcal{L}_\text{cls}$). The experimental results are shown in Table~\ref{tab:ablation}. 
After removing the classification loss, we observe that the average performance of the model decreased by 0.49\% and 0.52\% in terms of F1 score and EM score, respectively. 
While removing the regularization term, there was a significant performance decrease in handling \textit{irrelevant retrieval noise}. 


\begin{figure}[h]
  \centering
  \includegraphics[width=\linewidth]{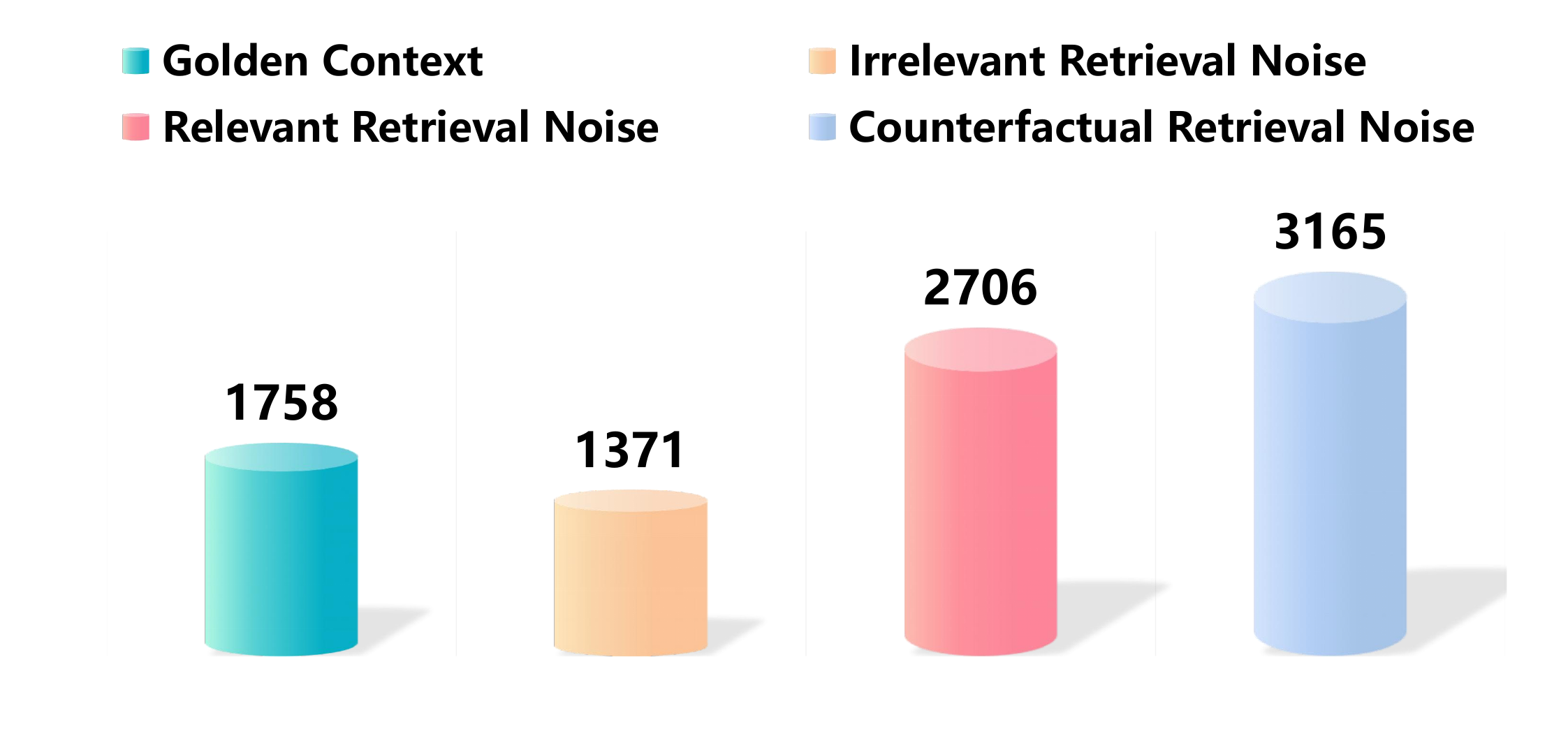}
  \caption{The number of queries and parameter updates are 4,500 and 9,000, respectively. The statistical content in this table pertains to different types of retrieval noises selected by RAAT each time the model parameters undergo an update.
  }
  \label{fig:number}
\end{figure}


\subsection{Further Discussion}
\paragraph{What types of adversarial samples does RAAT employ during training?}
To gain a comprehensive understanding of the underlying mechanisms of RAAT, particularly its utilization of specific retrieved data to augment model robustness, we undertook an in-depth examination of its training process, involving meticulously tracking the training iterations and conducting a thorough statistical analysis to quantify the number of different types of adversarial examples incorporated during the training phase.
The statistical results are illustrated in Figure~\ref{fig:number}. 
We observe that RAAT prioritizes the selection of adversarial examples that can significantly improve model robustness, as reflected in its tendency to choose certain types of adversarial examples. This is consistent with our empirical findings described in Section 3.2. RAAT tends to select adversarial examples associated with relevant retrieval noise and counterfactual retrieval noise for training. 

\section{Conclusion}
This work initially investigated retrieval noises in RALMs and categorized them into three distinct types, reflecting real-world environments. In addition, we introduced RAAT as a solution to address the noise robustness challenges faced by RALMs, which leveraged adaptive adversarial learning and multi-task learning to enhance the model's capability. Moreover, we established a benchmark to verify the effectiveness of RAAT based on three open-domain QA datasets. Experimental results demonstrate substantial improvements in F1 and EM scores for the LLaMA2 7B model fine-tuned with RAAT across diverse noise conditions.

\section{Limitations}
In this section, we delve into the limitations inherent in our work, with the objective of pinpointing areas for refinement and bolstering the performance of our model in future endeavors. Two principal limitations have been identified.
Firstly, the benchmark constructed for our experiments relies exclusively on datasets sourced from three open-domain question answering repositories. Going forward, we intend to compile additional high-quality datasets from varying NLP tasks and endeavor to retrieve texts from a more extensive array of knowledge bases. This strategic expansion aims to facilitate the creation of a more diversified and expansive benchmark tailored for evaluating the retrieval noise robustness of large language models.
Secondly, within the framework of RAAT, our efforts have been singularly concentrated on fortifying the retrieval noise robustness at the LLM end. However, the prospect of jointly training large language models and retrieval models emerges as a promising avenue for enhancing the overall robustness of RALMs. Although this dimension was not the primary focal point of our current work, in our subsequent investigations into retrieval noise robustness, we plan to delve into this avenue. This approach would facilitate the synchronized progress of both the large language model and the retrieval model, contributing to an overall improvement in their robustness.

\section*{Acknowledgments}
Min Yang was supported by National Key Research and Development Program of China (2022YFF0902100), National Natural Science Foundation of China (62376262), the Natural Science Foundation of Guangdong Province of China (2024A1515030166), Shenzhen Science and Technology Innovation Program (KQTD20190929172835662), Shenzhen Basic Research Foundation (JCYJ20210324115614039). This work was supported by Alibaba Group through Alibaba Innovative Research Program, Postdoctoral Fellowship Program of CPSF (GZC20232873), GuangDong Basic and Applied Basic Research Foundation (2023A1515110718 and 2024A1515012003).

\bibliography{custom}
\clearpage

\appendix


\begin{figure}[h]
  \centering
  \includegraphics[width=\linewidth]{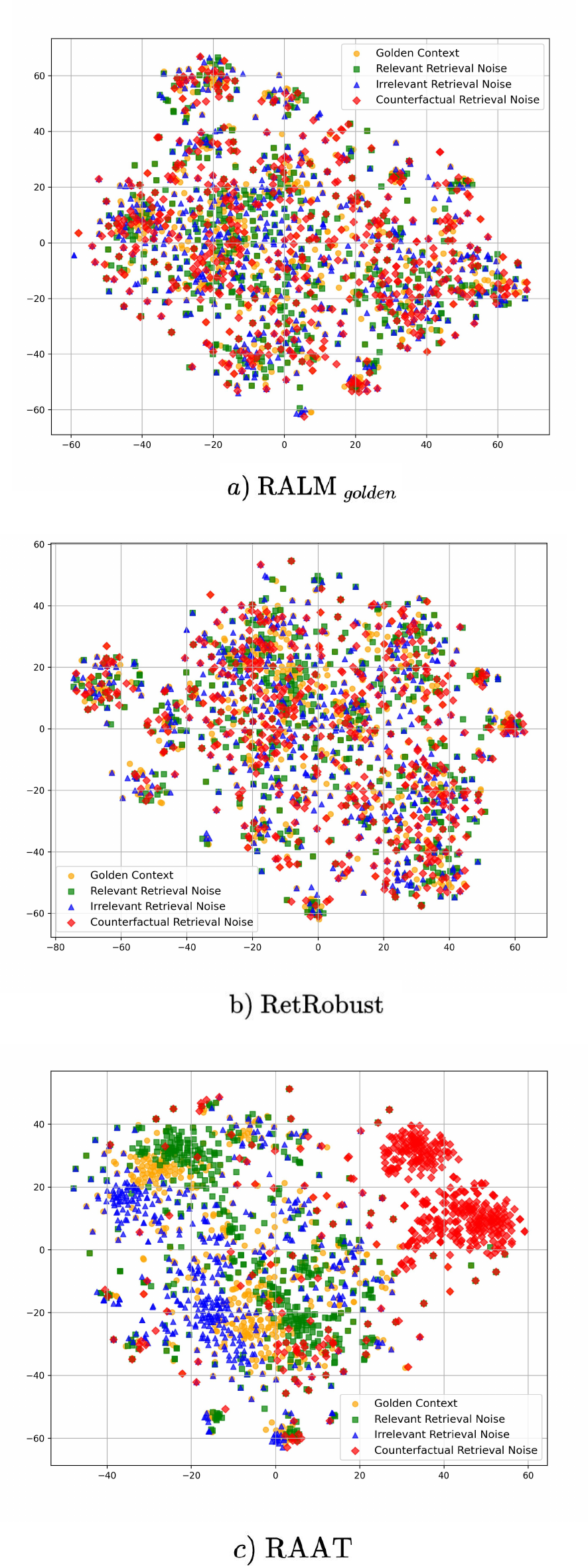}
  \caption{The results of T-SNE visualization. Following the introduction of four types of adversarial samples (i.e., retrieval noises) into models tuned by various methods, the hidden state of the last token is extracted. Subsequently, dimensionality reduction using t-SNE, clustering, and visualization are performed. This visual representation includes three methods, namely $\text{RALM}_{golden}$, $\text{RetRobust}$, and $\text{RAAT}$.}
  \label{fig:cluster}
\end{figure}

\section{Has the Model Truly Attained Noise Awareness?}
\label{sec:appendix}

Our preliminary investigation focused on the intrinsic capability of \textbf{RALM$_{golden}$} and \textbf{RetRobust} to classify the types of retrieval noises. Drawing inspiration from previous work~\cite{gueta2023knowledge}, we approached this matter through the application of clustering algorithms. The results, illustrated in Figure ~\ref{fig:cluster}, reveal suboptimal clustering of text vectors from \textbf{RALM$_{golden}$} and \textbf{RetRobust}, suggesting that the internal representations for noise classification in these models may lack clarity. Consequently, we introduced a noise classification loss $\mathcal{L}_{{cls}}$ into our RAAT method. The experimental results demonstrated tangible benefits with the incorporation of the classification loss. Additionally, we assessed the clustering effectiveness in models fine-tuned with RAAT, observing minimal distances among samples of irrelevant, relevant, and no retrieval noises, in contrast to the considerable distance from counterfactual retrieval noise samples. In particular, counterfactual retrieval noise posed the most significant challenge to LLMs; however, after RAAT tuning, it exhibited superior clustering and representation learning outcomes, indirectly validating the efficacy of RAAT.


\end{document}